\pdfoutput=1

\documentclass[a4paper, 10pt, conference]{ieeeconf}

\IEEEoverridecommandlockouts

\usepackage[pdftex]{graphicx}

\usepackage[colorinlistoftodos]{todonotes}
\usepackage{multirow}
\usepackage{hyperref}
\hypersetup{
    colorlinks=true,
    linkcolor=black,
    citecolor=black,
    filecolor=black,
    urlcolor=black,
}
\usepackage[T1]{fontenc}  
\usepackage[utf8]{inputenc}
\usepackage[english]{babel}
\usepackage[export]{adjustbox}
\usepackage[font=small]{caption}\usepackage{subcaption}
\usepackage[colorinlistoftodos]{todonotes}

\usepackage{amsmath, amssymb, amsfonts, amsthm, fouriernc}
\usepackage{mathtools}

\usepackage{csquotes}

\title{\LARGE \bf
Towards Cooperative Motion Planning \\ for Automated Vehicles in Mixed Traffic
}

\author{Maximilian Naumann$^{1}$ and Christoph Stiller$^{1,2}$%
\thanks{*We gratefully acknowledge support of this work by the Tech Center a-drive}%
\thanks{$^{1}$The authors are with FZI Research Center for Information Technology,
        Mobile Perception Systems, 76131 Karlsruhe, Germany
        {\tt\small naumann@fzi.de}}%
\thanks{$^{2}$The author is also with Karlsruhe Institute of Technology (KIT),
		Institute of Measurement and Control, 76131 Karlsruhe, Germany
        {\tt\small stiller@kit.edu}}%
}

\begin{document}

\maketitle
\thispagestyle{empty}
\pagestyle{empty}

\begin{abstract}

While motion planning techniques for automated vehicles in a reactive and anticipatory manner are already widely presented, approaches to cooperative motion planning are still remaining.
In this paper, we present an approach to enhance common motion planning algorithms, that allows for cooperation with human-driven vehicles.
Unlike previous approaches, we integrate the prediction of other traffic participants into the motion planning, such that the influence of the ego vehicle's behavior on the other traffic participants can be taken into account.
For this purpose, a new cost functional is presented, containing the cost for all relevant traffic participants in the scene. 
Finally, we propose a path-velocity-decomposing sampling-based implementation of our approach for selected scenarios, which is evaluated in a simulation.

\end{abstract}

\section{Introduction}
\label{introduction}

In the field of intelligent vehicles, tremendous progress has been achieved in the last decades \cite{BenglerDietmayerFarberMaurerStillerWinner2014ITSM}. 
With the first successful experiments of close-to-production cars in real traffic \cite{ZieglerAl2014ITSMag}, automated driving has gained more and more attention in public. 

In order to improve the reliability and thus the safety of automated vehicles, but also to increase their efficiency, \textit{cooperation} is focused on in recent research.
Here, cooperation through explicit communication of (fused) sensor information and desired driving behaviour \cite{Englund2016GCDC} as well as negotiation of possible solutions \cite{Carlino2013AuctionBased, Elhenawy2015AuctionBased, Rewald2016AuctionBased} or centralized approaches \cite{Manzinger2017centralized} are frequently addressed.

However, as reported in \cite{Naumann2016Herausforderungen}, cooperative behavior does not require V2X-communication.
Furthermore, as automated vehicles will share the road with human-driven cars at least at the beginning, cooperation with human drivers in non-V2X-equipped cars is essential.
Also, a natural, cooperative, human-like behavior of automated vehicles potentially increases their social acceptance.

\begin{figure}

\begin{subfigure}{\columnwidth}
\includegraphics[width=\linewidth]{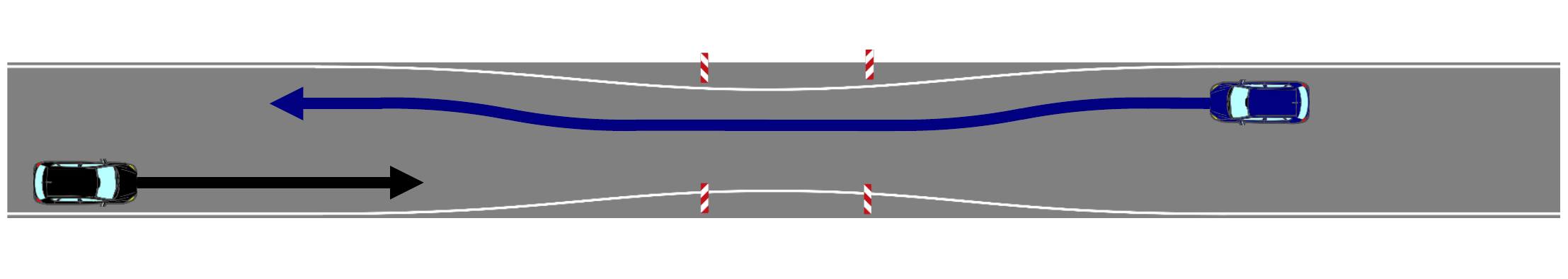} 
\caption{without signposted right of way}
\end{subfigure}

\begin{subfigure}{\columnwidth}
\includegraphics[width=\linewidth]{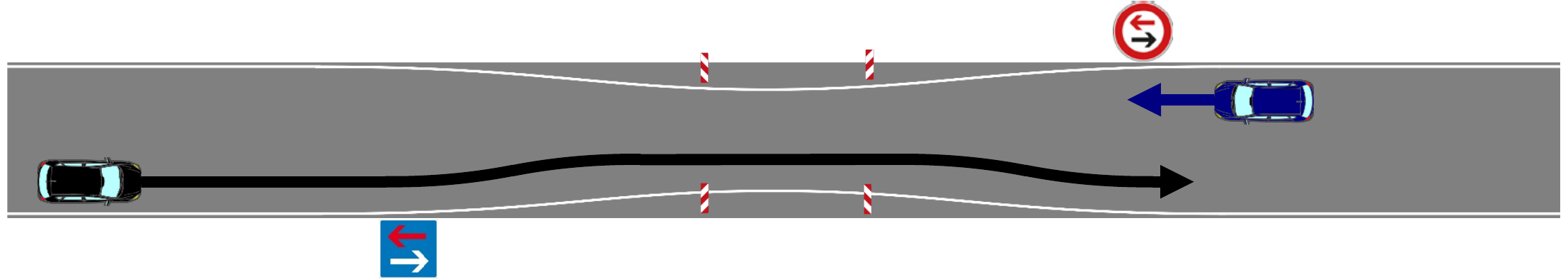}
\caption{with signposted right of way}
\end{subfigure}
 
\caption{Narrowing, with and without signposted right of way.}
\label{fig:narrowing_behavior}
\end{figure}

To the best of the authors' knowledge, previous motion planning approaches treated other traffic participants as obstacles which are to be avoided, similar to static obstacles like parked cars \cite{ZieglerAl2014ITSMag, Gonzales2016MotionPlanningSurvey}. 
While such approaches can deal with many everyday situations, such as driving autonomously or following other vehicles, some maneuvers, such as overtaking with oncoming traffic or passing a narrowing (cf. Figure  \ref{fig:narrowing_behavior}), require combinatorial approaches, as already reported by \cite{ZieglerAl2014ITSMag}. 
Still, even with combinatorial considerations as proposed by \cite{Bender2015combinatorial}, \cite{Qian2016MIQP}, cooperative behavior cannot be implemented:
If the motion prediction of other traffic participants is done isolated from the motion planning for the ego-vehicle, the behavior can be foresighted, but not cooperative in a bidirectional manner \cite{Naumann2016Herausforderungen}.
According to a study about German road traffic, cooperative behavior on average only occurs in the scale of one cooperative action per hour per traffic participant \cite{Benmimoun2004Effizienzsteigerung}.
Thus, their treatment by a separate method, besides the conventional motion planning, is reasonable.

This paper addresses the problem of cooperative motion planning without V2X-communication.
We propose a cost functional for trajectory ensembles, consisting of one trajectory per vehicle.
Thereby, we acknowledge the fact that not only the behavior of other traffic participants affects us, but also our behavior affects the others in a closed loop.
We consider the motion planning problem as the problem to find a globally optimal solution for a specific situation, knowing that every traffic participant has a different viewpoint considering optimality.
The costs depend on vehicle dynamics, passenger comfort, driving intention and trajectory clearance, as well as the traffic regulations, as further outlined in Section \ref{sec:global_opt_approach}.
In this approach, the prediction of other traffic participants is integrated into the motion planning.
As the assumption of cooperative behavior might be violated by some traffic participants, this risk is assessed and the trajectory is only driven if a safe "plan B" \cite{Damerow2015RiskAversive} trajectory is still possible in case of unexpected behavior.
An implementation of this approach is presented in Section \ref{sec:implementation}.
The  proposed  algorithm  is  finally  evaluated  in  Section \ref{sec:evaluation}.

\section{Global Optimum Approach}
\label{sec:global_opt_approach}
This section introduces the main building blocks of our approach to cooperative motion planning.
Central to this approach is the assumption that all traffic participants are aware of each other and therefore react on each other's behavior in a closed loop.
Subsequently, the trajectories for all relevant traffic participants are considered as one trajectory ensemble, and the quality of the solution depends on the trajectory of every participant separately as well as on the pairwise relation of the trajectories among each other.

This section is structured as follows: First, the representation of one trajectory in the ensemble is introduced. Subsequently, the cost functional is introduced. Next, before a solution is selected, the limitations to this approach are treated by a "plan B".

\subsection{Behavior Policy}
\label{sec:behavior_policy}
Cooperative motion planning is aware of the interaction of traffic participants.
Therefore, wrong assumptions concerning the behavior of other traffic participants might cause undesired behavior.
Even though, theoretically, any feasible behavior is possible, the authors make the following assumption: \textit{Every traffic participant follows the traffic regulations, as long as this compliant behavior is physically feasible.}

Consequently, assuming perfect perception, a collision involving our vehicle can only be caused by violating the traffic regulation without foreseeable reason while our reaction at the time of violation is insufficient to avoid the collision.

Arising from this assumption, we pursue the following policies:
\begin{itemize}
  \item If we have to give way, we can exclude a collision independent of others' behavior.
  \item If we have the right of way or the situation is not clearly regulated, we can exclude a collision if others behave rule compliant.
\end{itemize}

\subsection{Trajectory Representation}
For the representation of a single, deterministic trajectory, the established method of \cite{ZieglerAl2014TrajPlanning} is chosen:
The trajectory $\mathbf{x}(t) = (x(t), y(t))^T$ is a mapping $\mathbb{R} \rightarrow \mathbb{R}^2$, with tangent angle
$\psi$ and curvature $\kappa$.
A trajectory ensemble consists of one trajectory per traffic participant:  $\mathbf{X} = (\mathbf{x}^1, \mathbf{x}^2, ...)$, where the superscript describes the participants identifier.

\subsection{Cost Functional}
\label{sec:approach:cost_functional}
As proposed in \cite{Naumann2016Herausforderungen}, the quality of a solution, given by a trajectory ensemble, is determined by a cost functional.
The lowest costs denote the best solution.
Costs exceeding a certain value represent an infeasible solution.
The cost functional is the sum of the costs of every traffic participant $i$
\begin{equation*}
	G_{\mathrm{total}} = \sum_{i} G_i .
\end{equation*}

The costs $G_i$ pursue two main goals: They ensure the feasibility of the trajectory but also rate its comfort and effectiveness for a single car.
For this reason, the properties of the trajectory, such as velocity and acceleration, are rated with multiple evaluation functionals:

The feasibility costs exceed a certain bound if a trajectory is physically not feasible.
The pleasantness costs reflect the wish of the passenger to travel steady and comfortable, including the perceived safety of the journey.
Furthermore, the costs should motivate compliance with the traffic regulations. %
Not yielding is avoided by upscaling the costs of the vehicle that has the right of way in the pairwise trajectory costs.

In this approach, the ability to cooperate is associated with the ability to estimate the cost or quality of a solution for other traffic participants.

With the above information, the costs $G_i$ per participant can be split into costs $G_{i, 0}$ that only concern the own trajectory and costs $G_{i, j}$ that consider the relation to other trajectories:

\begin{equation*}
	G_{i} = G_{i, 0} + \sum_{j} G_{i, j}
\end{equation*}

\subsubsection{Formulation of the trajectory properties}
Analog to \cite{ZieglerAl2014TrajPlanning} the properties of the trajectory that are examined by the evaluation functionals are
\begin{itemize}
  \item the velocity $\mathbf{v}(t) = \dot{\mathbf{x}}(t)$
  \item the acceleration $\mathbf{a}(t) = \ddot{\mathbf{x}}(t)$
  \item the jerk $\mathbf{j}(t) = \dddot{\mathbf{x}}(t)$
  \item the distance to the left and right driving \\ corridor bound $d_{\mathrm{left}} ( \mathbf{x}(t))$ and $d_{\mathrm{right}} ( \mathbf{x}(t))$
  \item the yaw rate $\omega(t) = \dot{\psi}(t)$ and
  \item the curvature $\kappa (t)$ .
\end{itemize}

Additionally, properties of trajectory pairs describe their distance to each other.
The shortest spatial distance
is described by
\begin{equation*}
	d_{\mathrm{min}}(\mathbf{x}^1(t), \mathbf{x}^2(t)) = \min_t \left( d(\mathbf{x}^1(t), \mathbf{x}^2(t), t) \right),
\end{equation*}
where $d$ denotes a distance measure between states of different vehicles.

\begin{figure}
\centering
\begin{subfigure}{0.49\linewidth}
\centering
\includegraphics[width=0.6\linewidth]{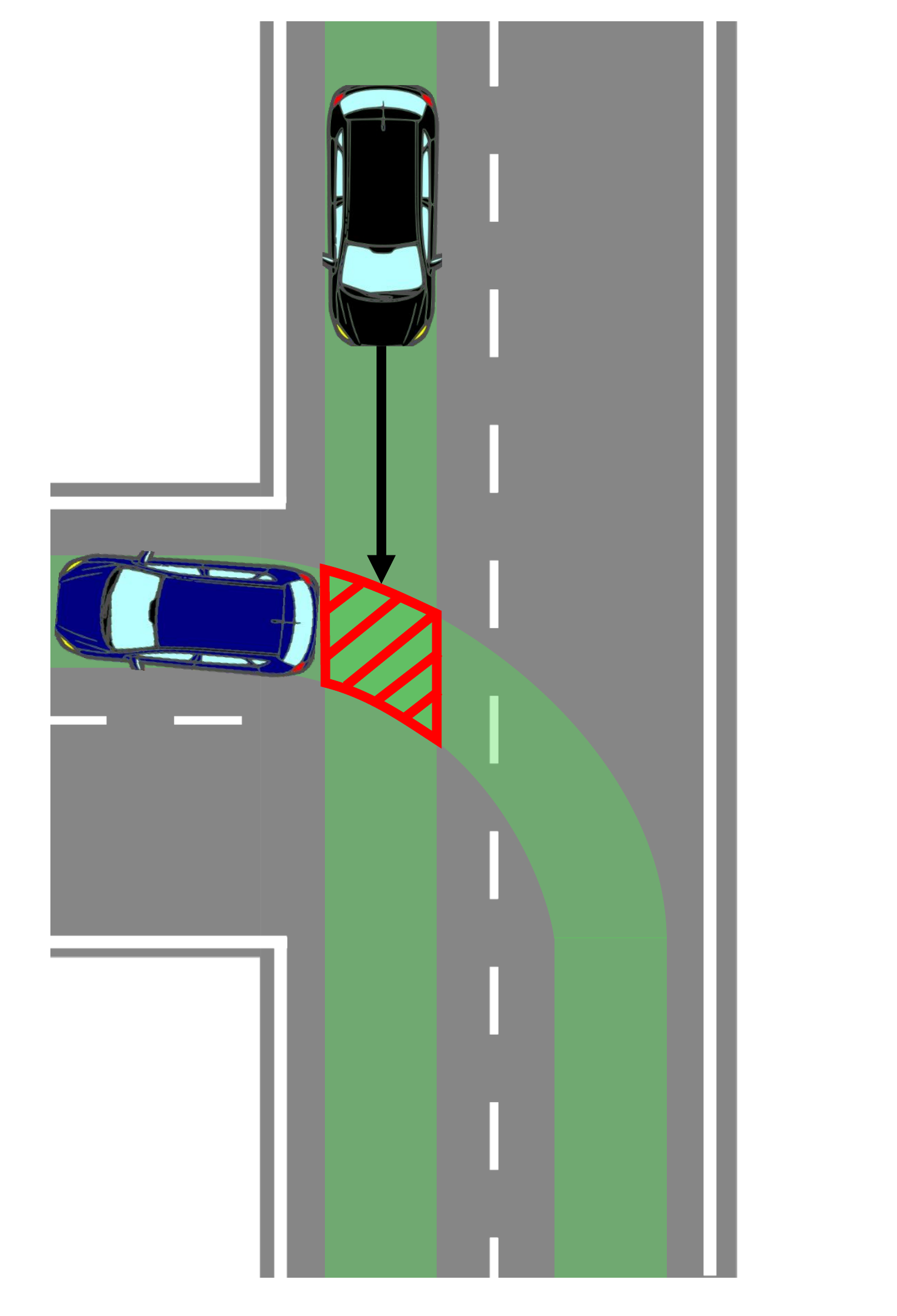}
\caption{blue vehicle drove first
}
\end{subfigure}
\begin{subfigure}{0.49\linewidth}
\centering
\includegraphics[width=0.6\linewidth]{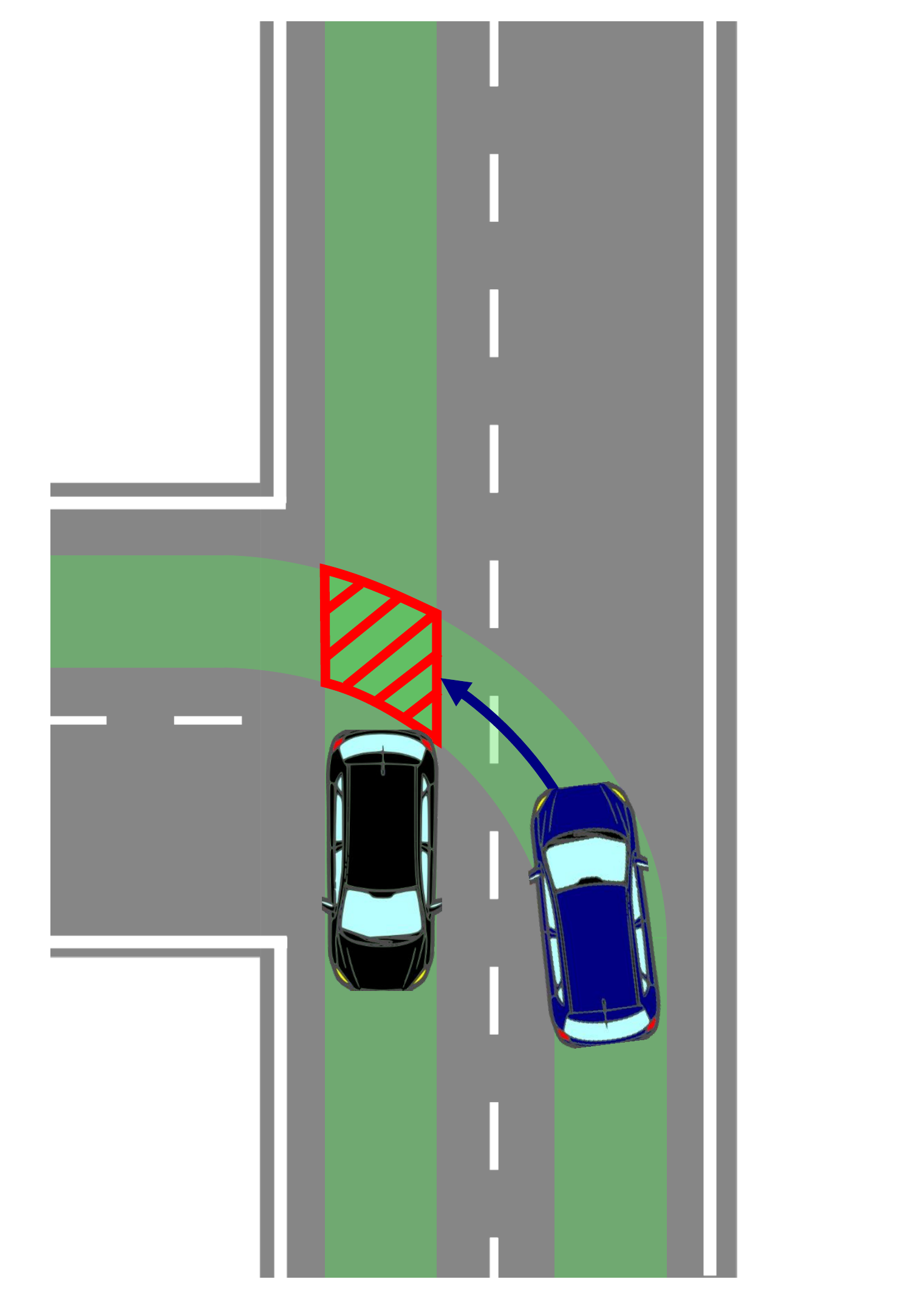}
\caption{black vehicle drove first
}
\end{subfigure}
\caption{The TZC is the time that the second vehicle takes to enter the red potential collision zone, assuming constant velocity.}
\label{fig:TZC}
\end{figure}

To account for the perceived safety, but also to obey the traffic regulations, another property is introduced.
Here, we can make use of time-referenced measures, as they equal a velocity-referenced spatial distance measure.
In general, a collision is only possible if paths overlap.
When determining the criticality, respectively the collision risk, of two trajectories, their closest point in time and space is crucial.
Regarding a violation of the right of way, Cooper investigated the \textit{post encroachment time (PET)} for specific scenarios \cite{Cooper1984PET}.
Based on the latter, also regarding the potential collision zone, we propose the \textit{time of zone clearance (TZC)} as a measure for the criticality of two trajectories with overlapping paths:
The TZC is the time that elapses between the first vehicle leaving potential collision zone and the second vehicle entering this area, independent of the right of way (cf. Figure \ref{fig:TZC}).

Given the paths are overlapping and given the trajectories are not colliding, the TZC is calculated as follows:
\begin{eqnarray*}
	\mathrm{TZC}
	&=&
	\mathrm{TZC} (\mathbf{x}^\mathrm{first}(t), \mathbf{x}^\mathrm{second}(t)) \\
	&=&
	\frac{\text{gap along path}}{\text{velocity of the second vehicle}} \\
	&=&
	\frac{s^{\mathrm{second}} (t_{\mathrm{second,in}}) - s^{\mathrm{second}} (t_{\mathrm{first,out}})}
	{v^\mathrm{second}(t_\mathrm{first,out})} \\
\end{eqnarray*}
with $s^{\mathrm{second}}$ being the path of the vehicle that passes the collision zone second, $v^{\mathrm{second}}$ being the scalar velocity along this path, $t_\mathrm{first,out}$ being the time at which the first vehicle clears the collision zone and $t_{\mathrm{second,in}}$ being the time at which the second vehicle enters the collision zone.
Constant velocity is chosen as passengers cannot foresee the planned trajectory and as it reflects possible actions (maximum deceleration or acceleration) best.

If the paths do not overlap, the $\mathrm{TZC}$ is defined to be infinite, if the trajectories collide, it is less or equal zero.

\subsubsection{Formulation of the evaluation functionals}
As in this work the costs are also calculated for human-driven cars in order to predict their behavioral decisions, they should reflect humans' understanding of the quality of a trajectory.
Therefore, the previously introduced scalar trajectory properties $f(\mathbf{X})$ are investigated.
Vectorial properties, such as the acceleration, are therefore split into their longitudinal and lateral part, using a motion model.

The costs of a trajectory are subdivided into three zones:
\begin{itemize}
	\item comfort zone $\mathcal{Z}_\mathrm{comf}$
	\item discomfort zone $\mathcal{Z}_\mathrm{disc}$
	\item infeasibility zone $\mathcal{Z}_\mathrm{inf}$
\end{itemize}
each for positive ($^+$) and negative ($^-$) deviation from the optimum $f_\mathrm{opt}$.
The functionals $G(f)$ expressing the costs induced by a trajectory property $f$ are called \textit{evaluation functionals}.

For the sake of steadiness and piecewise differentiability, all costs are starting from zero at their lower bound but do not vanish at the start of the next zone.
Accordingly, the total costs $G$ are defined as
\begin{equation*}
   G(f) =
   \begin{cases}
     G_\mathrm{comf} & \text{, } f \in \mathcal{Z}_\mathrm{comf} \\
     G_\mathrm{comf} + G_\mathrm{disc} & \text{, } f \in \mathcal{Z}_\mathrm{disc} \\
     G_\mathrm{comf} + G_\mathrm{disc} + G_\mathrm{inf} & \text{, } f \in \mathcal{Z}_\mathrm{inf} .
   \end{cases}
\end{equation*}

The comfort component induces only little costs
\begin{equation*}
	G_\mathrm{comf}^{+} \left( f \right) = a_\mathrm{+} \cdot \left(\Delta f_\mathrm{comfort}^\mathrm{+}\right)^2
\end{equation*}
depending on the distance of $f$ to the optimal value
\begin{equation*}
    \Delta f_\mathrm{comf}^\mathrm{+} =
	\Delta f_\mathrm{comf}^\mathrm{+}(\mathbf{X}) =
	f(\mathbf{X}) - f_\mathrm{opt}.
\end{equation*}

Consequently, given a comfort threshold $T_\mathrm{comf}$ and assuming a comfortable deviation $\Delta f_\mathrm{cmargin}^\mathrm{+}$, the parameter $a_\mathrm{+}$ is to be set to
\begin{equation*}
    a_\mathrm{+}
    =
    \frac{T_\mathrm{comf}}
    {\left(\Delta f_\mathrm{cmargin}^\mathrm{+}\right)^2} .
\end{equation*}
The costs $G_\mathrm{comf}^{-}$ for comfortable negative deviation are calculated correspondingly with the parameter $a_\mathrm{-}$.

The discomfort costs rise quadratic, but direction-dependent:
\begin{equation*}
	G_\mathrm{disc}^{+} \left( f \right) = b_\mathrm{+} \cdot \left(\Delta f_\mathrm{disc}^\mathrm{+}\right)^2
\end{equation*}
depending on the distance of $f$ to the upper start of the discomfort zone $f_\mathrm{disc, start}^\mathrm{+}$
\begin{equation*}
    \Delta f_\mathrm{disc}^\mathrm{+} =
	\Delta f_\mathrm{disc}^\mathrm{+}(\mathbf{X}) =
	f(\mathbf{X}) - f_\mathrm{disc}^\mathrm{+}.
\end{equation*}
For logical reasons, the parameter $b_\mathrm{+}$ should be notably higher than $a_\mathrm{+}$.
Negative deviations are treated correspondingly with the parameter $b_\mathrm{-}$.

Before the property represents the infeasibility of a trajectory, the infeasibility costs rise exponentially
\begin{equation*}
	G_\mathrm{inf}^{+} \left( f \right) = c_\mathrm{+} \cdot \left(\Delta f_\mathrm{inf}^\mathrm{+}\right)^2
	\cdot e^{\lvert \Delta f_\mathrm{inf}^\mathrm{+} \rvert }
\end{equation*}
depending on the distance of $f$ to the upper infeasible value $f_\mathrm{inf}^\mathrm{+}$ minus a margin $f_\mathrm{margin}^\mathrm{+}$ from which the costs start rising
\begin{equation*}
    \Delta f_\mathrm{inf}^\mathrm{+} =
	\Delta f_\mathrm{inf}^\mathrm{+}(\mathbf{X}) =
	f(\mathbf{X}) - \left(f_\mathrm{inf}^\mathrm{+} - \Delta f_\mathrm{margin}^\mathrm{+}\right) .
\end{equation*}
Consequently, given an infeasibility threshold $T_\mathrm{inf}$ and assuming a margin $\Delta f_\mathrm{imargin}^\mathrm{+}$, the parameter $c_\mathrm{+}$ is to be set to
\begin{equation*}
    c_\mathrm{+}
    =
    \frac{T_\mathrm{inf}}
    {\left(\Delta f_\mathrm{imargin}^\mathrm{+}\right)^2 \cdot e^{\lvert \Delta f_\mathrm{imargin}^\mathrm{+} \rvert }} .
\end{equation*}
Further, the infeasibility zone $\mathcal{Z}_\mathrm{inf}$ includes the margin in this notation.
Again, negative deviations are treated correspondingly with the parameter $c_\mathrm{-}$.

\subsubsection{Formulation of the cost functional}
\begin{figure}
  \begin{center}
  \includegraphics[width=0.89\linewidth]{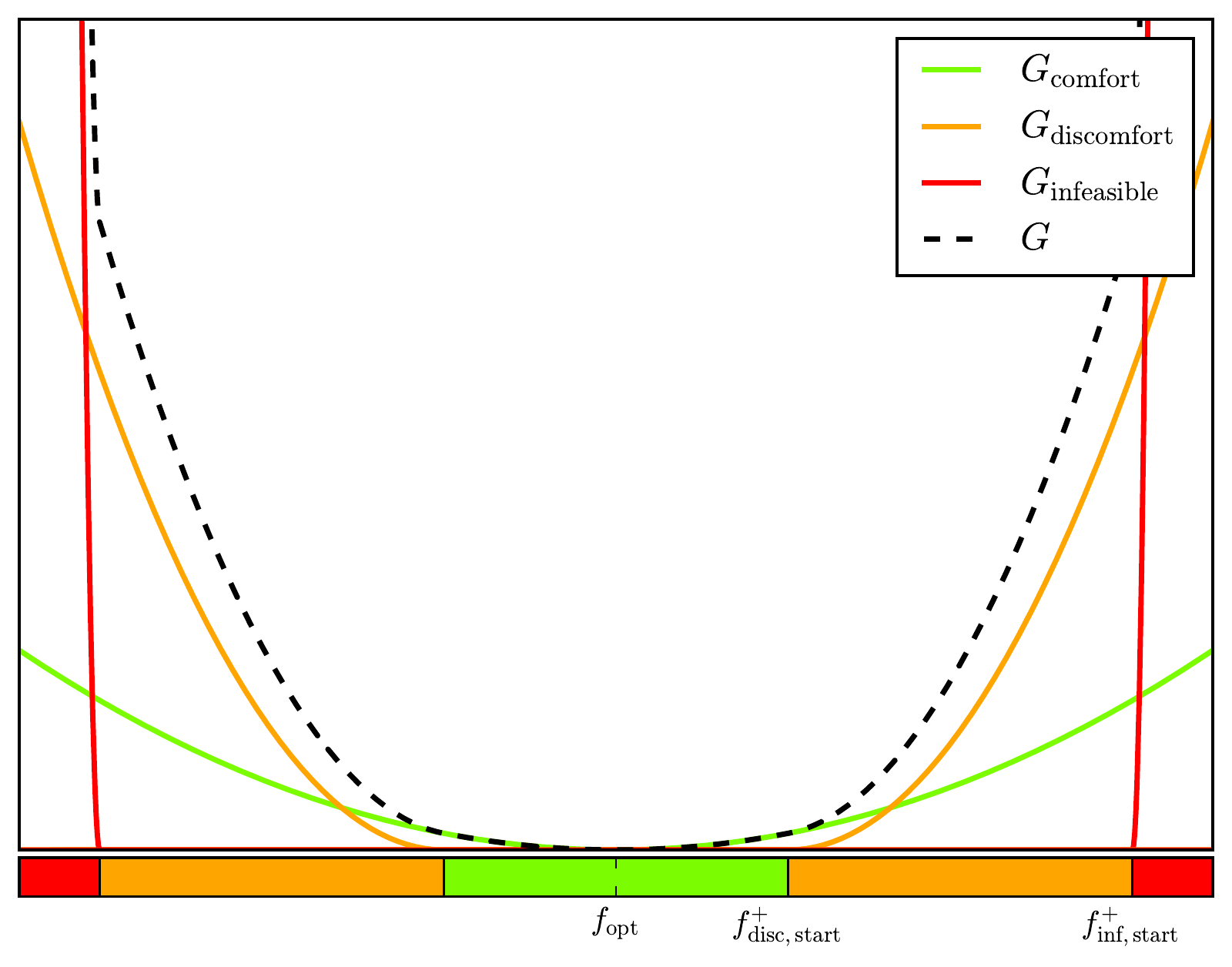}
  \end{center}
\caption{Composition of the cost function $G$ for a single trajectory property $f$: Very low costs around the optimum value $f_\mathrm{opt}$, increasing rapidly in close vicinity of $f_\mathrm{inf}$.}
\label{fig:plot_eval_functions}
\end{figure}

With the evaluation functionals, the cost functional for a single property $f$ is composed as follows (cf. Figure \ref{fig:plot_eval_functions}):
\begin{eqnarray*}
    G \left( f \right)
    =&&
    G_\mathrm{comf}^\mathrm{+} \cdot \sigma(\Delta f_\mathrm{comf}^\mathrm{+})
    +
    G_\mathrm{comf}^\mathrm{-} \cdot \sigma(-\Delta f_\mathrm{comf}^\mathrm{-})\\
    &+&
    G_\mathrm{disc}^\mathrm{+} \cdot \sigma(\Delta f_\mathrm{disc}^\mathrm{+})
    +
    G_\mathrm{disc}^\mathrm{-} \cdot \sigma(-\Delta f_\mathrm{disc}^\mathrm{-})\\
    &+&
    G_\mathrm{inf}^\mathrm{+} \cdot \sigma(\Delta f_\mathrm{inf}^\mathrm{+})
    +
    G_\mathrm{inf}^\mathrm{-} \cdot \sigma(-\Delta f_\mathrm{inf}^\mathrm{-}), 
\end{eqnarray*}
where $\sigma$ denotes the step function.
The right of way of $i$ over $j$ is acknowledged by adding the comfort-related costs of vehicle $i$, upscaled with factor $u$, to the pairwise trajectory costs, if $i$ has the right of way:
\begin{equation*}
	G_{i,j,\mathrm{row}}
	=
	u \left( G_{i, 0,\mathrm{comf}} + G_{i, 0,\mathrm{disc}} \right) .
\end{equation*}
A suitable choice of $u$ ensures that the right of way is heeded, but its violation is still feasible, as stated in Section \ref{sec:behavior_policy}.

The full cost functional is composed as follows:
 \begin{equation*}
	G_{\mathrm{total}}(\mathbf{X})
	=
	\sum_{i} \left( G_{i, 0}(\mathbf{x}^i) + \sum_{j} G_{i, j}(\mathbf{x}^i, \mathbf{x}^j) \right)
\end{equation*}
with singleton trajectory costs for vehicle~$i$

\begin{equation*}
    G_{i, 0}(\mathbf{x}^i)
    =
    G_{\mathbf{v}} + G_{\mathbf{a}} +
    G_{\mathbf{j}} + G_{\omega} +
    G_{\kappa} + G_{\mathrm{offset}}
\end{equation*}
and pairwise trajectory costs for vehicle~$i$ due to vehicle~$j$
\begin{equation*}
    G_{i, j}(\mathbf{x}^i, \mathbf{x}^j)
    =
    G_{\mathrm{TZC}} + G_{d_\mathrm{min}} + G_{\mathrm{row}} .
\end{equation*}

\subsection{Plan B}
In order to obey our previously introduced policy, plan~B trajectories are to be checked, as proposed in \cite{Damerow2015RiskAversive}.
By doing so, we avoid maneuvering into situations that lead to collisions, if we made wrong assumptions concerning the behavior of other traffic participants.
As their execution is unlikely, we accept discomfortable but feasible trajectories.
This corresponds to a neglection of the comfort terms in the upper cost functional.
As with the previous trajectories, plan~B trajectories can be calculated via a local continuous method \cite{ZieglerAl2014TrajPlanning}, a sampling-based method such as RRT$^{*}$ \cite{Karaman2011rrt_star} or other approaches.

\subsection{Selection of Solution}
\label{sec:approach:solution_selection}
As for passenger comfort, the evaluation of the TZC should already cause high discomfort costs at around $2\:s$, a security margin is induced intrinsically by this approach.
Thus, even a very small optimum, represented by a small range of minimal costs, does not equal a physically optimal trajectory, that would pass objects as close as possible in space-time.
Rather, it already contains those security margins that are considered comfortable by humans and that consequently should be feasible with measurement uncertainties in the range of human perception errors.
Hence, the optimum point can be chosen independent of its wideness, as long as a valid plan~B protects the approach against consequences of wrong assumptions.

\section{Implementation}
\label{sec:implementation}
In the following, a first approach for cooperative motion planning in specific situations, based on the previously introduced cost functional, is presented.

\subsection{Path-Velocity Decomposition}

Several potentially cooperative situations have highly constraint driving corridors for the traffic participants, independent of the order and number of traffic participants.
Consequently, we make use of the \textit{path-velocity decomposition (PVD)}, as introduced by \cite{Kant1986PVD}.
The calculation of paths in static environments has already been widely investigated.
Hence, valid paths are considered predefined (cf. Figure \ref{fig:left_turn_with_and_without_priority}) and the implementation focuses on the velocity profiles along the paths.

\subsection{Sampling}

As the optimization problem is non-convex, but the control variable for the velocity of each vehicle is only one-dimensional, a classical sampling approach is chosen.
Therefore, the trajectories $\mathbf{x}(t)$ are approximated by discretization in equidistant time steps: %
\begin{equation*}
	\mathbf{x}_i = \mathbf{x}(t_i), \; \; \; \; t_i = t_0 + i \Delta t .
\end{equation*}

For each car, multiple trajectories are sampled:
Starting with an initial position and velocity, a random jerk sequence determines the velocity profiles and thus the trajectory.
Next, the overall costs of each trajectory ensemble are calculated.
For the solutions with the lowest costs, the plan~B trajectory is checked until a valid plan~B is found.

\subsection{Plan B}
Instead of using a different planning method with the assumption or classical prediction of disadvantageous behavior of others, we again make use of the PVD:
Given the paths, a collision is only possible in particular areas that can be determined a priori.
Thus, unlike in \cite{Damerow2015RiskAversive}, no trajectory has to be planned.
Rather, the plan~B-consideration can be seen as a \enquote{what could I do if}-consideration.
The key questions are:
In every time step, what could the other vehicle do that leads to a collision with us?
And what could we do to avoid this?
This consideration can be split into the following cases:

\subsubsection{Other vehicle drives first}
If the other vehicle drives first, it can only cause a collision by deceleration.
In reaction, we can decelerate as well.
If we can manage to stop before the collision zone, we have a valid plan~B.

\subsubsection{Ego vehicle drives first}
If the ego vehicle drives first, the other vehicle can only cause a collision by acceleration.
In reaction, we can also accelerate, to still drive first, or decelerate to stop before the other vehicle collides with us.
As the path is regarded as predefined, changing the path is not considered.

\subsection{Implications on the cost functional}

Since in this implementation, trajectories are discretized in time, derivatives are approximated by finite differences.
Thus, the functionals of section \ref{sec:approach:cost_functional} turn into functions.
Consequently, trajectory properties that depend on a single minimum, such as TZC, can be largely affected if this minimum is not sampled.
In order to avoid this, either the sampling rate must be sufficiently high, or the point of the exact minimum has to be interpolated.
As this implementation is not based on linear optimization but on sampling, we interpolate the crucial points.

The jerk is not considered to avoid high order derivatives.
Also, the curvature itself is not considered as the predefined path guarantees the compliance with the steering geometry.
However, it is used to calculate the lateral acceleration values.
Furthermore, the  shortest spatial distance $d_\mathrm{min}$ either lies in the collision zone and is considered by the TZC, or it is not relevant.
Hence, it is neglected as well.

\subsection{Selection of Solutions}
As explained in section \ref{sec:approach:solution_selection}, criticality protection is ensured via the costs of the TZC and the check for a plan~B.
Consequently, the solution with the lowest costs and a valid plan~B is selected and its ego trajectory is executed, as long as its costs do not exceed the feasibility-threshold.
In case no solution has a valid plan~B, an emergency braking maneuver is triggered.
Note: In in-car applications, the parallel running classical, reactive motion-planner would have to take over control in this case.

\section{Evaluation}
\label{sec:evaluation}

\begin{figure}
\centering
\begin{subfigure}{0.49\linewidth}
\centering
\includegraphics[width=0.6\linewidth]{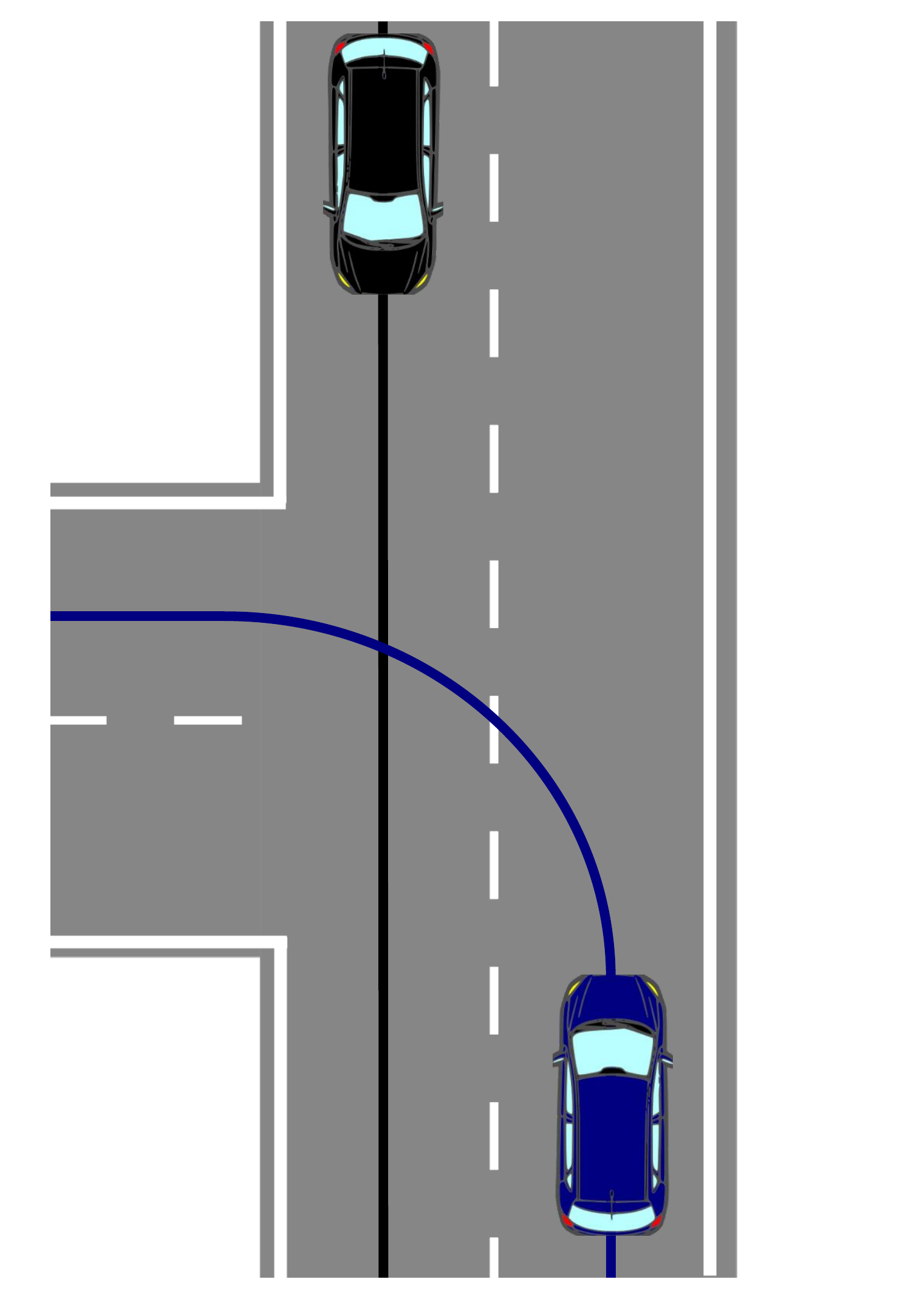} 
\caption{%
}
\end{subfigure}
\begin{subfigure}{0.49\linewidth}
\centering
\includegraphics[width=0.6\linewidth]{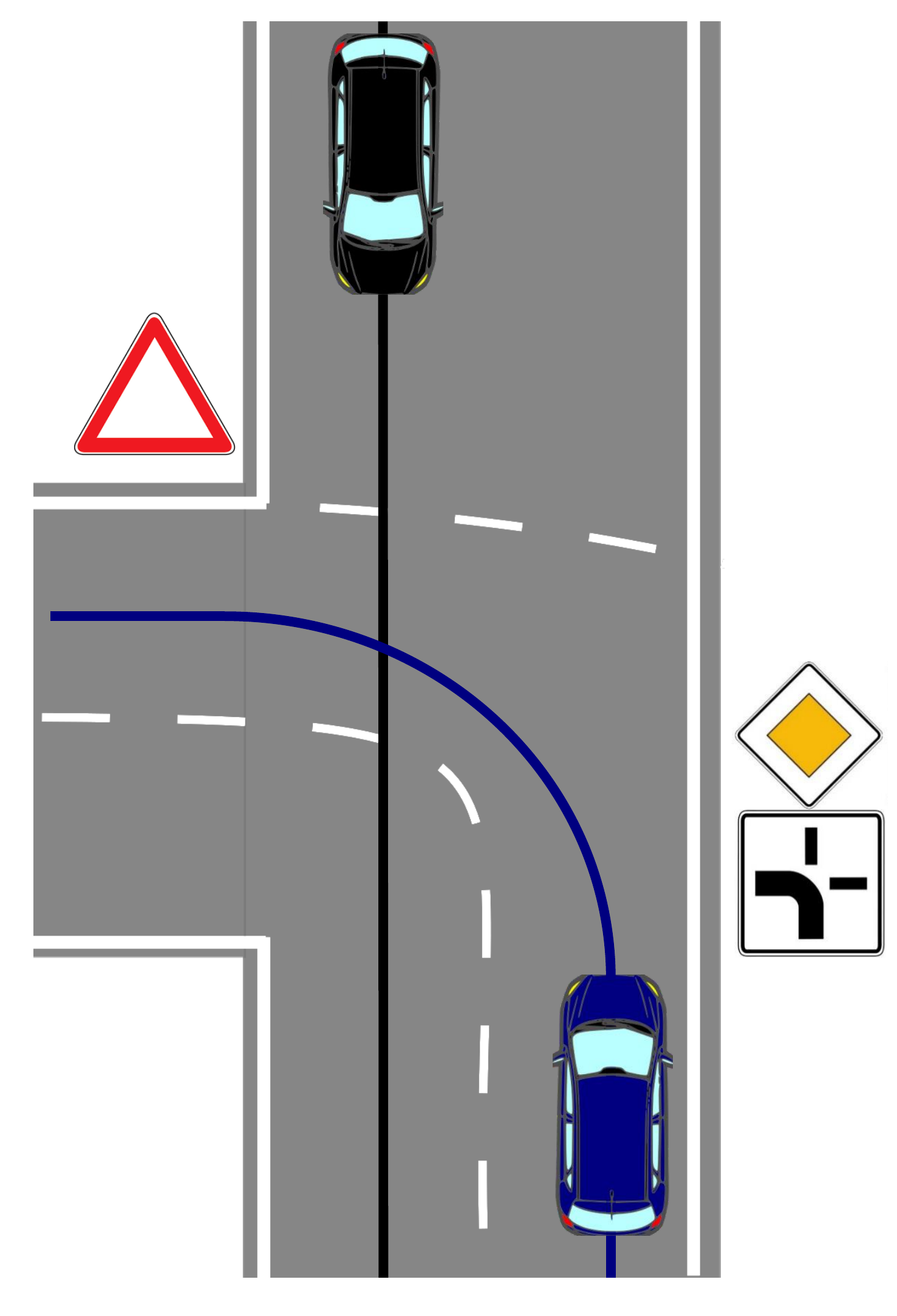}
\caption{%
}
\end{subfigure}
\caption{Left turn at T-junction, with and without signposted right of way and predefined paths.}
\label{fig:left_turn_with_and_without_priority}
\end{figure}

In this section, the method outlined in section \ref{sec:implementation} is evaluated for two scenarios, a left turn at a T-intersection 
and passing through a narrowing of the road 
(cf. Figures \ref{fig:narrowing_behavior} and \ref{fig:left_turn_with_and_without_priority}).

\subsection{Simulation}

For both scenarios, each with and without signposted right of way, but sharing the same paths, velocity profiles were sampled.
From the resulting trajectories, ensembles with one trajectory per vehicle were generated.
In order to reduce computational cost, trajectories that did not reach the end of the collision zone were excluded from the cost calculation.
Furthermore, colliding trajectory ensembles were excluded.
The remaining ensembles were analyzed with respect to
\begin{itemize}
	\item comfort costs
	\item discomfort costs
	\item infeasibility costs
	\item traffic regulation costs.
\end{itemize}

\subsection{Analysis}

\begin{figure}
  \begin{center}
  \includegraphics[width=0.92\linewidth]{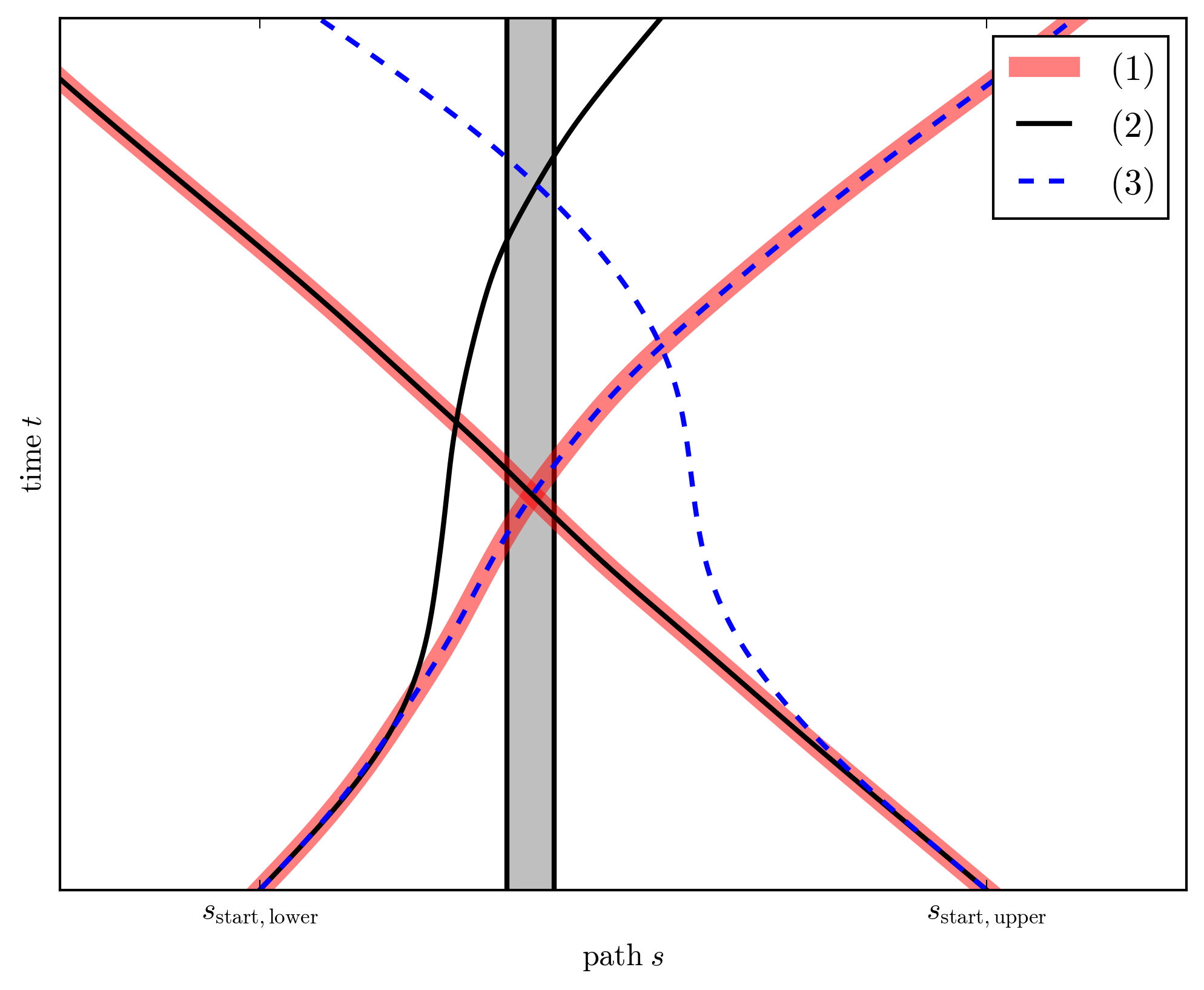}
  \end{center}
\caption{Minimum cost trajectories in the T-junction scenario with the collision zone marked in grey: (1) for each vehicle solely on the road, (2) when upper vehicle has right of way (Fig.~\ref{fig:left_turn_with_and_without_priority}a), (3) when lower vehicle has right of way (Fig.~\ref{fig:left_turn_with_and_without_priority}b). }
\label{fig:s_t_leftturn}
\end{figure}

\begin{figure}
  \begin{center}
  \includegraphics[width=0.92\linewidth]{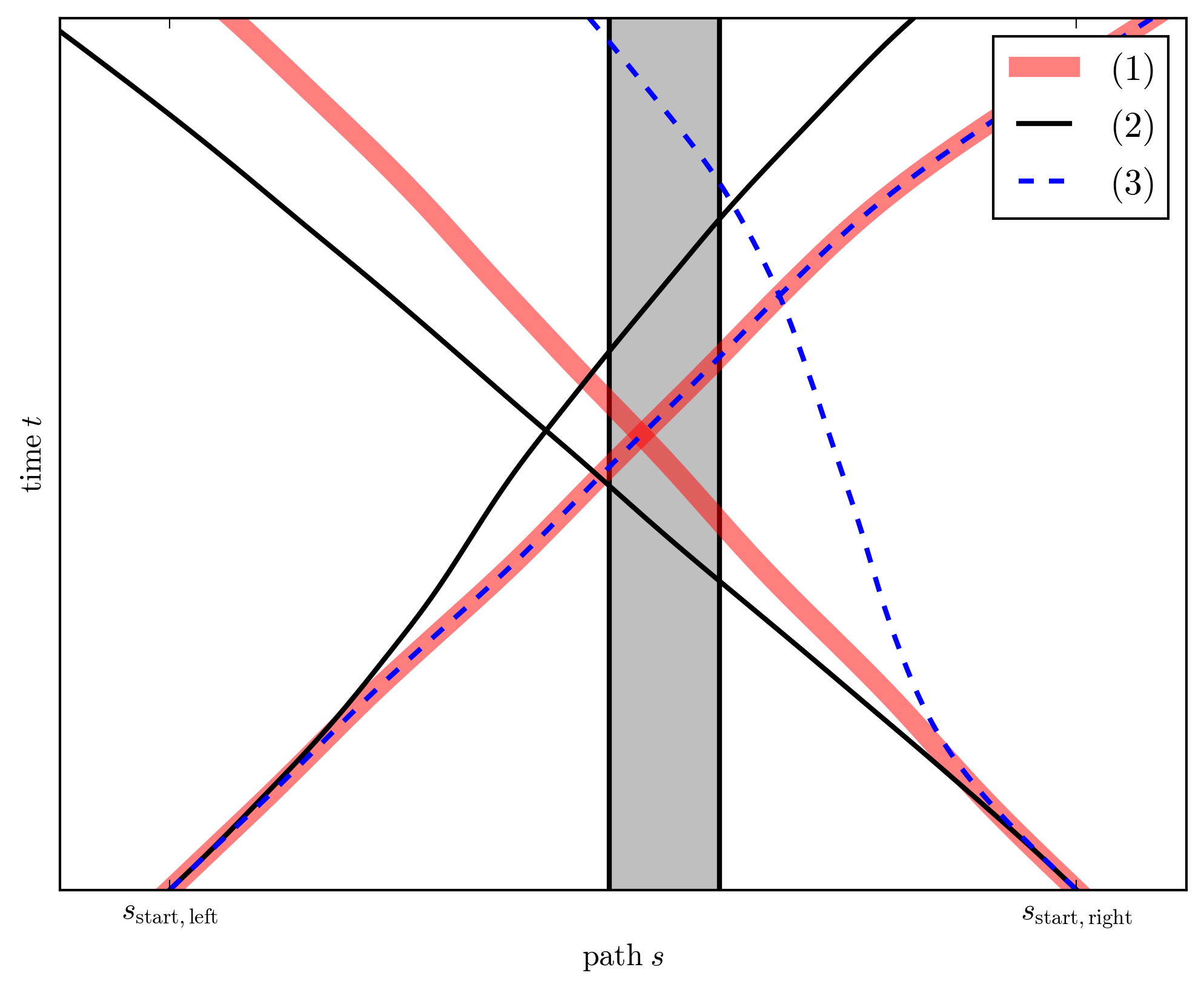}
  \end{center}
\caption{Minimum cost trajectories in the narrowing scenario with the collision zone marked in grey: (1) for each vehicle solely on the road, (2) when no right of way predefined (Fig.~\ref{fig:narrowing_behavior}a), (3) when left vehicle has right of way (Fig.~\ref{fig:narrowing_behavior}b). }
\label{fig:s_t_narrowing}
\end{figure}

As depicted in Figure \ref{fig:s_t_leftturn} and \ref{fig:s_t_narrowing}, the initial states were chosen in a way that the optima of both vehicles overlap in the collision zone. 
In the T-junction scenario, the right of way is regulated with and without traffic signs. 
A violation of the right of way causes high costs so the optimal solution is following the rules. The trajectory of the vehicle that has right of way is not interfered (cf. Figure \ref{fig:s_t_leftturn} (2) and (3)).

In the narrowing scenario, the right of way is not regulated without traffic signs.
Here, due to equal cost parameters, the vehicle that is closer to the narrowing passes first.
Still, traffic signs can overrule this globally most comfortable solution and shift the optimum (cf. Figure \ref{fig:narrowing_behavior} and \ref{fig:s_t_narrowing} (3)).

If a collision can only be avoided by one of the vehicles, as the other is too close to the collision zone, the optimal solution is the collision avoidance.
Even though this violates the traffic regulations, the infeasibility costs overrule discomfort costs and traffic regulation costs.

Further, if we do not interfere a vehicle that has the right of way, its costs $G_{\mathrm{row}}$ are constantly high, but not raised by our behavior.
In this case, the optimal solution is that we pass first, without violation of traffic rules.

\section{Conclusions and Future Work}
\label{sec:conclusion}

In this paper, we presented a new approach to cooperative motion planning, able to cooperate with human drivers and automated vehicles without requiring V2X-communication.
While the approach is valid for two-dimensional motion planning, our first implementation covers several scenarios deploying PVD.

The preliminary results for the simulated scenarios 
demonstrate that the method produces safe and comfortable cooperative trajectories in a narrowing and a typical intersection scenario.
Individual trajectory costs have been extended by costs accounting for mutual comfort and safety of any pair of trajectories.
Other traffic participants have been taken into account by incorporating their individual costs.
The total trajectory costs for each participant have been segmented into three areas representing comfortable driving, uncomfortable driving and collision/infeasibility.

Future work includes real time implementation and on-road experiments with our vehicle "BerthaOne".
Several parametrizations will be used for the cost functional, considering different vehicle types and driver behaviors.
Furthermore, probabilistic trajectories will be accommodated to account for inherent uncertainties in perception and behavior.  

\bibliographystyle{IEEEtran}
\bibliography{references}

\end{document}